\documentclass{article} 
\usepackage{nips15submit_e,times}
\usepackage{hyperref}
\usepackage{url}
\usepackage[numbers]{natbib}
\usepackage{wrapfig}
\usepackage{times}
\usepackage{graphicx}
\usepackage{amsmath}
\usepackage{amssymb}

\newcommand{\para}[1]{\noindent{\textbf{#1}}}

\newcommand{\fullname}[0]{Frequency-Sensitive Hashed Nets}
\newcommand{\abbrev}[0]{FreshNets}
\newcommand{\abbrevtable}[0]{FreshNets}

\newcommand{\bx}[0]{{\boldsymbol x}}
\newcommand{\by}[0]{{\boldsymbol y}}
\newcommand{\bA}[0]{{\boldsymbol A}}
\newcommand{\bB}[0]{{\boldsymbol B}}
\newcommand{\bW}[0]{{\boldsymbol W}}
\newcommand{\bV}[0]{{\boldsymbol V}}

\newcommand{\bw}[0]{{\boldsymbol w}}

\newcommand{\mR}[0]{\mathbb{R}}
\newcommand{\bN}[0]{\mathbb{N}}
\newcommand{\mL}[0]{\mathcal{L}}

\newcommand{\mV}[0]{\mathcal{V}}
\newcommand{\bmV}[0]{\boldsymbol{\mathcal{V}}}

\newcommand{\tV}[0]{V^{k\ell}} 
\newcommand{\btV}[0]{\bV^{k\ell}} 
\newcommand{\fV}[0]{\mV^{k\ell}} 
\newcommand{\bfV}[0]{{\bmV}^{k\ell}} 

\newcommand{\fdct}[0]{f_{dct}}
\newcommand{\fidct}[0]{f^{-\!1}_{dct}}

\newcommand{\grad}[2]{ \frac{\partial {#1}}{\partial #2} }

\newcommand{\naive}[0]{na\"{\i}ve}

\newcommand{\data}[1]{{\sc #1}}

\newcommand{\ie}[0]{\emph{i.e.}}
\newcommand{\wrt}[0]{\emph{w.r.t.}}
\newcommand{\etal}[0]{et al.}

\newcommand{\boldBlue}[1]{\textcolor{blue}{$\mathbf{#1}$}}

\title{Compressing Convolutional Neural Networks}

\author{
Wenlin Chen, ~James T. Wilson\\
Washington University in St. Louis\\
\texttt{\{wenlinchen, j.wilson\}@wustl.edu } \\
\And
Stephen Tyree\\
NVIDIA, Santa Clara, CA, USA\\
\texttt{styree@nvidia.com} \\
\AND
Kilian Q. Weinberger, ~Yixin Chen\\
Washington University in St. Louis \\
\texttt{kilian@wustl.edu, chen@cse.wustl.edu} \\
}

\nipsfinalcopy 

\begin{document}

\maketitle

\begin{abstract}
%
Convolutional neural networks (CNN) are increasingly used in many areas of computer vision. They are particularly attractive because of their ability to ``absorb'' great quantities of labeled data through millions of parameters. However, as model sizes increase, so do the storage and memory requirements of the classifiers. We present a novel network architecture, \fullname{} (\abbrev{}), which exploits inherent redundancy in both convolutional layers and fully-connected layers of a deep learning model, leading to dramatic savings in memory and storage consumption. Based on the key observation that the weights of learned convolutional filters are typically smooth and low-frequency, we first convert filter weights to the frequency domain with a discrete cosine transform (DCT) and use a low-cost hash function to randomly group frequency parameters into hash buckets. All parameters assigned the same hash bucket share a single value learned with standard back-propagation.
To further reduce model size we allocate fewer hash buckets to high-frequency components, which are generally less important.
We evaluate \abbrev{} on eight data sets, and show that it leads to drastically better compressed performance than several relevant baselines.

\end{abstract}


\section{Introduction}

In the recent years convolutional neural networks (CNN) have lead to impressive results in object recognition~\cite{krizhevsky2012imagenet}, face verification~\cite{schroff2015facenet} and audio classification~\cite{lee2009unsupervised}. Problems that seemed impossibly hard only five years ago can now be solved at better than human accuracy~\cite{he2015delving}. 
Although CNNs have been known for a quarter of a century~\cite{fukushima1980neocognitron}, only recently have their superb generalization abilities been accepted widely across the machine learning and computer vision communities.
This broad acceptance coincides with the release of very large collections of labeled data~\cite{deng2009imagenet}. Deep networks and CNNs are particularly well suited to learn from large quantities of data, in part because they can have arbitrarily many parameters. As data sets grow, so do model sizes. In 2012, the first winner of the ImageNet competition that used a CNN had already 240MB of parameters and the most recent winning model, in 2014, required 567MB~\cite{SimonyanZ14a}.

Independently, there has been another parallel shift of computing from servers and workstations to mobile platforms. As of January 2014 there have already been more web searches through smart phones than computers\footnote{\url{http://tinyurl.com/omd58sq}}. Today speech recognition is primarily used on cell phones with intelligent assistants such as Apple's Siri, Google Now or Microsoft's Cortana. As this trend continues,  we are expecting machine learning applications to also shift increasingly towards mobile devices. 
%
However, the disjunction of deep learning with ever increasing model sizes and mobile computing reveals an inherent dilemma. Mobile devices have tight memory and storage limitations. For example, even the most recent iPhone 6 only features 1GB of RAM, most of which must be used by the operating system or the application itself. In addition, developers must make their apps compatible with the most limited phone still in circulation, often restricting models to just a few megabytes of parameters.

In response, there has been a recent interest in reducing the model sizes of deep networks. Denil et al.~\cite{denil2013predicting} use low-rank decomposition of the weight matrices to reduce the effective number of parameters in the network. Buciluǎ et al. \cite{bucilua2006model} and Ba et al.~\cite{Caruana2014} show that complex models can be compressed into 1-layer neural networks. Independently, the model size of neural networks can be reduced effectively through reduced bit precision~\cite{courbariaux2015low}. 

In this paper we propose a novel approach for neural network compression targeted especially for CNNs. We build on recent work by Chen et al.~\cite{chen2015compressing}, who show that weights of fully connected networks can be effectively compressed with the hashing trick~\cite{weinberger09feature}.
Due to the nature of local pixel correlation in images (\ie{} spatial locality), filters in CNNs tend to be smooth. We transform these filters into frequency domain with the discrete cosine transform (DCT)~\cite{rao2014discrete}. In frequency space, the filters are naturally dominated by low frequency components. Our compression takes this smoothness property into account and randomly hashes the frequency components of all CNN filters at a given layer into one common set of hash buckets. All components inside one hash bucket share the same value. As lower frequency components are more pronounced than higher frequencies, we allow collisions only between similar frequencies and allocate fewer hash buckets for the high frequencies (which are less important). 

Our approach has several compelling properties: 
1. The number of parameters in the CNN is \emph{independent} of the number of convolutional filters; 2. During testing we only need to add a low-cost hash function and the inverse DCT transformation to any existing CNN code for filter reconstruction; 3. During training, the hashed weights can be learned with simple back-propagation~\cite{bishop1995neural}---the gradient of a hash bucket value is the sum of gradients of all hashed frequency components in that bucket.

We evaluate our compression scheme on eight deep learning image benchmark data sets and compare against four competitive baselines. Although all compression schemes lead to lower test accuracy as the compression increases, our \abbrev{} method is by far the most effective compression method and yields the lowest generalization error rates on almost all classification tasks.

\section{Background}
\label{sec:back}
\para{Feature Hashing} (\emph{a.k.a} the hashing trick)~\cite{dasgupta2010sparse,hashKernelShi:2009,weinberger09feature} has been previously studied as a technique for reducing model storage size. 
In general, it can be regarded as a dimensionality reduction method that maps an input vector $\bx\in\mR^d$ to a much smaller feature space via a mapping   $\phi\colon\!\mR^d\rightarrow \mR^k$ where $k\ll d$. The mapping $\phi$ is a composite of two approximately uniform auxiliary hash functions $h\colon\!\bN\!\rightarrow\!\{1,\dots,k\}$ and $\xi\colon\!\bN \!\rightarrow\!\{-1,+1\}$. The $j^{th}$ element of the $k$-dimensional hashed input is defined as
\begin{equation*}
	\phi_j(\bx)=\sum_{i:h(i)=j}\xi(i)\ x_i.
\end{equation*}
As shown in \cite{weinberger09feature}, a key property of feature hashing is its preservation of inner product operations, where inner products after hashing produce the correct pre-hash inner product in expectation:
\begin{equation*}
	\mathbb{E}[\phi(\bx)^\top\phi(\by)]_{\phi} = \bx^\top\by.
\end{equation*}
This property holds because of the bias correcting sign factor $\xi(i)$. 
With feature hashing, models are directly learned in the much smaller space $\mR^k$, which not only speeds up training and evaluation but also significantly conserves memory.
For example, a linear classifier in the original space could occupy $O(d)$ memory for model parameters, but when learned in the hashed space only requires $O(k)$ parameters.
The information loss induced by hash collision is much less severe for sparse feature vectors and can be counteracted through multiple hashing~\cite{hashKernelShi:2009} or larger hash tables~\cite{weinberger09feature}. 

\para{Discrete Cosine Transform (DCT)~\cite{rao2014discrete}.}\label{sec:dct}
Methods built on the DCT are widely used for compressing images and movies, including forming the standard technique for JPEG~\cite{wallace1991jpeg}.
DCT expresses a function as a weighted combination of sinusoids of different phases/frequencies where the weight of each sinusoid reflects the magnitude of the corresponding frequency in the input.
When employed with sufficient numerical precision and without quantization or other compression operations, the DCT and inverse DCT (projecting frequency inputs back to the spatial domain) are lossless.
Compression is made possible in images by local smoothness of pixels (\emph{e.g.} a blue sky) which can be well represented regionally by fewer non-zero frequency components.
Though highly related to the discrete Fourier transformation (DFT), DCT is often preferable for compression tasks because of its \emph{spectral compaction} property where weights for most images tend to be concentrated in a few low-frequency components of the DCT~\cite{rao2014discrete}.
Further, the DCT transformation yields a real-valued representation, unlike the DFT whose representation has imaginary components. 
Given an input matrix $\bV\!\in\! \mR^{d\times d}$, the corresponding matrix $\bmV\!\in\! \mR^{d\times d}$ in frequency domain after DCT is defined as:
\begin{equation}
	\mV_{j_1 j_2} = s_{j_1} s_{j_2} \sum_{i_1=0}^{d-1} \sum_{i_2=0}^{d-1} c(i_1,i_2,j_1,j_2)\ V_{i_1 i_2},
	\label{eq.dct}
\end{equation}
\begin{equation*}
    \textrm{where\ \ \ } c(i_1,i_2,j_1,j_2) =  \cos{\left[\frac{\pi}{d} \left(i_1+\frac{1}{2}\right)j_1 \right]} 
	\cos{\left[\frac{\pi}{d} \left(i_2+\frac{1}{2}\right)j_2 \right]}
\end{equation*}
is the cosine basis function, and
$s_{j}\!=\!\sqrt{\frac{1}{d}}$ when $j\!=\!0$ and $s_{j}\!=\!\sqrt{\frac{2}{d}}$ otherwise.
We use the shorthand $\fdct$ to denote the DCT operation in Eq. \eqref{eq.dct}, \ie{} $\bmV=\fdct(\bV)$. The inverse DCT converts $\bmV$ from the frequency domain back to the spatial domain, reconstructing $\bV$ without loss:
\begin{equation}
	V_{i_1 i_2} = \sum_{j_1=0}^{d-1}  \sum_{j_2=0}^{d-1} s_{j_1} s_{j_2}\ c(i_1,i_2,j_1,j_2)\ \mV_{j_1 j_2}.
	\label{eq.idct}
\end{equation}
We denote the inverse DCT function in Eq. \eqref{eq.idct} as $\fidct$, \ie{} $\bV = \fidct(\bmV)$.

\section{\fullname{}}


\begin{wrapfigure}{r}{0.4\textwidth}
\vspace{-3ex}
    \includegraphics[width=0.4\textwidth]{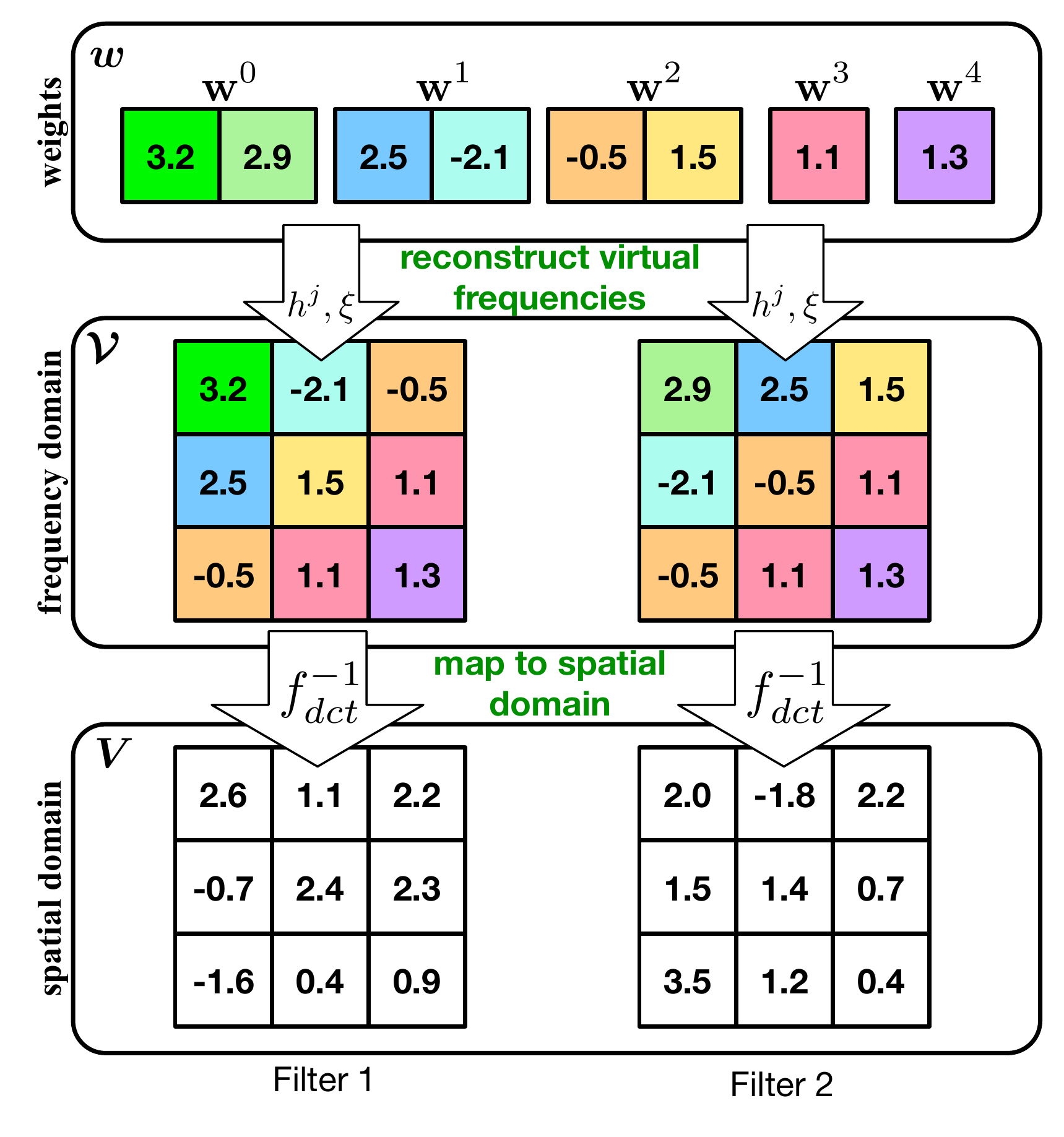} 
    \vspace{-4ex}
    \caption{A schematic illustration of \abbrev{}. Two spatial filters are re-constructed from the frequency weights in vector $\bw$. The frequency weights are accessed with two hash functions and then transformed to the spatial domain. The vector $\bw$ is partitioned into sub-vectors $\bw^j$ shared by all entries with similar frequency (corresponding to index sum $j=j_1+j_2$). Colors indicate which hash bucket was accessed.}
    \label{fig:dctHashedNets}
    \vspace{-4ex}
\end{wrapfigure}
Here we present \abbrev{}, a method for using weight sharing to reduce the model size (and memory demands) of convolutional neural networks.
Similar to the work of Chen et al.~\cite{chen2015compressing}, we achieve smaller models by randomly forcing weights throughout the network to share identical values.
Unlike previous work, we implement the weight sharing and gradient updates of convolutional filters in the \emph{frequency domain}.
These sharing constraints are made prior to training, and we learn frequency weights under the sharing assignments.
Since the assignments are made with a hash function, they incur no additional storage.


\para{Filters in spatial and frequency domain.} 
Let the matrix $\bV^{k\ell}\!\in\!\mR^{d\times d}$ denote the weight matrix of the $d\!\times\! d$ convolutional filter that connects the $k^{th}$ input plane to the $\ell^{th}$ output plane.
(For notational convenience we assume square filters and only consider the filters in a single layer of the network.)
The weights of all filters in a convolutional layer can be denoted by a $4$-dimensional tensor $\bV\!\in\!\mR^{m\times n\times d \times d}$ where $m$ and $n$ are the number of input planes and output planes, respectively, resulting in a total of $m\times n\times d^2$ parameters.
Convolutional filters can be represented equivalently in either the spatial or frequency domain, mapping between the two via the DCT and its inverse.
We denote the filter in frequency domain as $\bmV^{k\ell} \!=\! \fdct(\bV^{k\ell})\!\in\!\mR^{d\times d}$ and recover the original spatial representation through $\btV = \fidct(\bfV)$, as defined in Eq. \eqref{eq.dct} and \eqref{eq.idct}, respectively.
The tensor of all filters is denoted $\bmV\!\in\!\mR^{m\times n\times d \times d}$.



\para{Random Weight Sharing by Hashing.}
We would like to reduce the number of model parameters to exactly $K$ values stored in a weight vector $\bw \!\in\!\mR^{K}$, where $K \ll m\times n\times d^2$.
To achieve this, we randomly assign a value from $\bw$ to each filter frequency weight in $\bmV$.
A \naive{} implementation of this random weight sharing would introduce an auxiliary matrix for $\bmV$ to track the weight assignments, using to significant additional memory.
To address this problem, Chen \etal{} \cite{chen2015compressing} advocate use of the hashing trick to (pseudo-)randomly assign shared parameters.
Using the hashing trick, we tie each filter weight $\fV_{j_1 j_2}$ to an element of $\bw$ indexed by the output of a hash function $h(\cdot)$:
\begin{equation}
	\fV_{j_1,j_2} = \xi(k, \ell, j_1,j_2)\ w_{h(k, \ell, j_1,j_2 )},
	\label{eq.hash_xi}
\end{equation}
where $h(k, \ell, j_1,j_2)\!\in\!\{1,\cdots,K\}$, and $\xi(k, \ell, j_1,j_2)\!\in\!\{\pm 1\}$ is a sign factor computed by a second hash function $\xi(\cdot)$ to preserve inner-products in expectation as described in Section~\ref{sec:back}.
With the mapping in Eq. \eqref{eq.hash_xi}, we can implement shared parameter assignments with no additional storage cost. (For a schematic illustration, see Figure~\ref{fig:dctHashedNets}. The figure also incorporates a frequency sensitive hashing scheme discussed later in this section.)

\para{Gradients over Shared Frequency Weights.}
Typical convolutional neural networks learn filters in the spatial domain. As our shared weights are stored in the frequency domain, we derive the gradient with respect to filter parameters in frequency space.
Following Eq. \eqref{eq.idct}, we express the gradient of parameters in the spatial domain \wrt{} their counterparts in the frequency domain:
\begin{equation}
	\grad{\tV_{i_1 i_2}}{\fV_{j_1 j_2}} = s_{j_1} s_{j_2}\ c(i_1,i_2,j_1,j_2).
	\label{eq.grad_w_w}
\end{equation}

Let $\mL$ be the loss function adopted for training.
Using standard back-propagation, we can derive the gradient \wrt{} filter parameters in the spatial domain, $\grad{\mL}{\tV_{i_1 i_2}}$.
By the chain rule with Eq. \eqref{eq.grad_w_w}, we express the gradient of $\mL$ in the frequency domain:
\begin{equation}
	\grad{\mL}{\fV_{j_1 j_2}} = \sum_{i_1=0}^{d-1} \sum_{i_2=0}^{d-1}
								 \grad{\mL}{\tV_{i_1 i_2}}\ \grad{\tV_{i_1 i_2}}{\fV_{j_1 j_2}} 
							  = s_{j_1} s_{j_2} \sum_{i_1=0}^{d-1} \sum_{i_2=0}^{d-1}
							  	 c(i_1,i_2,j_1,j_2)\ \grad{\mL}{\tV_{i_1 i_2}}.
\label{eq.grad}\end{equation}
Comparing with Eq. \eqref{eq.dct}, we see that the gradient in the frequency domain is merely the DCT of the gradient in the spatial domain:
\begin{equation}
	\grad{\mL}{\bfV} = \fdct{} \left(\grad{\mL}{\btV}\right).
	\label{eq.grad_update}
\end{equation} 


We compute gradient for each shared weight $w_h$ by simply summing over the gradient at each filter parameter where the weight is assigned, \ie{} all $\fV_{j_1 j_2}$ where $h=h(k,\ell,j_1,j_2)$:
\begin{align}
	\grad{\mL}{w_h}
			\ =\ \sum_{k=0}^{m} \sum_{\ell=0}^{n} \sum_{j_1=0}^{d-1} \sum_{j_2=0}^{d-1}
						\grad{\mL}{\fV_{j_1 j_2}} \grad{\fV_{j_1 j_2}}{w_h} 
			\ \ \ = \!\!\!\!\!\!\sum_{\substack{k,\ell,j_1,j_2: \\ h=h(k, \ell, j_1,j_2 )}}
						\!\!\!\!\!\!\!\!\!\xi(k, \ell, j_1,j_2)
						\left[ \fdct{} \left(\grad{\mL}{\btV}\right) \right]_{j_1 j_2}
\label{eq.full_grad}\end{align}
where $[\bA]_{j_1 j_2}$ denotes the $(j_1,j_2)$ entry in matrix $\bA$.

\begin{wrapfigure}{r}{0.30\textwidth}
\vspace{-4ex}
    \includegraphics[width=0.30\textwidth]{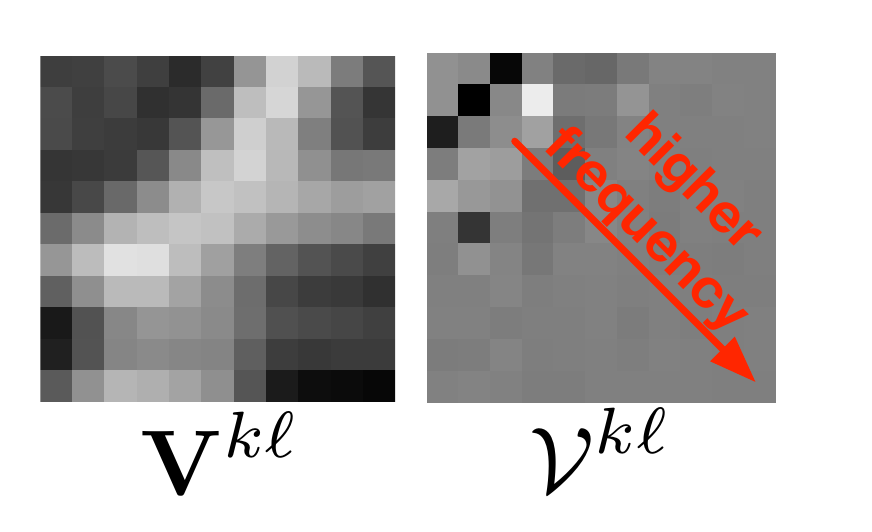} 
\vspace{-5ex}
    \caption{An example of a filter in spatial (left) and frequency domain (right). }
    \label{fig:vcurlyv}
    \vspace{-1ex}
\end{wrapfigure}
\para{Frequency Sensitive Hashing.}
\label{subsec.hierarchical_hashing}
Figure~\ref{fig:vcurlyv} shows a filter in spatial (left) and frequency (right) domains.
In the spatial domain CNN filters are smooth~\cite{krizhevsky2012imagenet} due to the local pixel smoothness in natural images.
In the frequency domain this corresponds to components with large magnitudes in the low frequencies, depicted in the upper left half of $\bfV$ in Figure~\ref{fig:vcurlyv}.
Correspondingly, the high frequencies, in the bottom right half of $\bfV$,
have magnitudes near zero.

As components of different frequency groups tend to be of different magnitudes (and thereby varying importance to the spatial structure of the filter), we want to avoid collisions between high and low frequency components.
Therefore, we assign separate hash spaces to different frequency groups.
In particular, 
we partition the $K$ values of $\bw$ into sub-vectors $\bw^0,\dots,\bw^{2d-2}$ of sizes $K_0,\dots,K_{2d-2}$, where $\sum_j K_j=K$.
This partitioning allows parameters with the same frequency, corresponding to their index sum $j=j_1+j_2$, to be hashed into a corresponding dedicated hash space $\bw^j$.
We rewrite Eq. \eqref{eq.hash_xi} with the new frequency sensitive shared weight assignments:
\begin{equation*}
	\fV_{j_1,j_2} = \xi(k, \ell, j_1,j_2)\ w^{j}_{h^{j}(k, \ell, j_1,j_2 )}
\end{equation*}
where $h^j(\cdot)$ maps an input key to a natural number in $\{1,\cdots,K_j\}$ and $j\!=\!j_1\!+\!j_2$.

We define a compression rate  $r_j\!\in\!(0,1]$ for each frequency region $j$ and assign $K_j\!=\!r_j N_j$.
A smaller $r_j$ induces more collisions during hashing, leading to increased weight sharing.
Since lower frequency components tend to be of higher importance, making collisions more hurtful, we commonly assign larger $r_j$ (fewer collisions) to low-frequency regions.
Intuitively, given a size budget for the whole convolutional layer, we want to squeeze the hash space of high frequency region to save space for low frequency regions.
These compression rates can either be assigned by hand or determined programmatically by cross-validation, as demonstrated in Section \ref{sec:exp}. 



\section{Related Work}
Several recent studies have confirmed that there is significant redundancy in the parameters learned in deep neural networks.
Recent work by Denil~\etal{}~\cite{denil2013predicting} learns parameters in fully-connected layers after decomposition into two low-rank matrices, \ie{} $\bW\!=\!\bA\bB$ where $\bW\!\in\!\mR^{m\times n}$, $\bA\!\in\!\mR^{m\times k}$ and $\bB\in \mR^{k\times n}$.
In this way, the original $O(m n)$ parameters could be stored with $O(k(m+n))$ storage, where $k\ll \min(m,n)$.
Several works apply related approaches to speed up the evaluation time with convolutional neural networks.
Two works propose to approximate convolutional filters by a weighted linear combination of basis filters~\cite{rigamonti2013learning,jaderberg2014speeding}.
In this setting, the convolution operation only needs to be performed with the small set of basis filters.
The desired output feature maps are computed by matrix multiplication as the weighted sum of these basis convolutions.
Further speedup can be achieved by learning rank-one basis filters so that the convolution operations are very cheap to compute~\cite{denton2014exploiting,lebedev2014speeding}.
Based on this idea, Denton \emph{et al.}~\cite{denton2014exploiting} advocate decomposing the four-dimensional tensor of the filter weights into a sum of different rank-one, four-dimensional tensors.
In addition, they adopt bi-clustering to group filters such that each subgroup can be better approximated by rank-one tensors.

In each of these works, evaluation time is the main focus, with any resulting storage reduction achieved merely as a side effect.
Other works focus entirely on compressing the fully-connected layers of CNNs~\cite{gong2014compressing,yang2014deep}.
However, with the trend toward architectures with fewer fully connected layers and additional convolutional layers \cite{szegedy2014going}, compression of filters is of increased importance.
Another technique for speeding up convolutional neural network evaluation is computing convolutions in the Fourier frequency domain, as convolution in the spatial domain is equivalent to (comparatively lower-cost) element-wise multiplication in the frequency domain~\cite{mathieu2013,vasilache2014fast}.
Unlike \abbrev{}, for a filter of size $d\times d$ and an image of size $n\times n$ where $n>d$, Mathieu~\etal{}~\cite{mathieu2013} convert the filter to its frequency domain of size $n\times n$ by oversampling the frequencies, which is necessary for doing element-wise multiplication with a larger image but also increases the memory overhead at test time.
Training in the Fourier frequency domain may be advantageous for similar reasons, particularly when convolutions are being performed over large 3-D volumes~\cite{brosch2015efficient}.

Most relevant to this work is HashedNets \cite{chen2015compressing} which compresses the fully connected layers of deep neural networks.
This method uses the hashing trick to efficiently implement parameter sharing prior to learning, achieving notable compression with less loss of accuracy than the competing baselines which relied on low-rank decomposition or learning in randomly sparse architectures.

\section{Experimental Results}\label{sec:exp}
In this section, we conduct several comprehensive experiments on benchmark datasets to evaluate the performance of \abbrev{}.

\paragraph{Datasets.}
We experiment with eight benchmark datasets: \data{cifar10}, \data{cifar100}, \data{svhn} and five challenging variants of \data{mnist}.
The \data{cifar10} dataset contains $60000$ images of $32\times32$ pixels with three color channels.
Images are selected from ten classes with each class consisting of $6000$ unique instances.
The \data{cifar100} dataset also contains $60000$ $32\times32$ images, but is more challenging since the images are selected from $100$ classes (each class has 600 images). 
For both {\sc cifar} datasets, $50000$ images are designated for training and the remaining $10000$ images for testing. 
To improve accuracy on \data{cifar100}, we augment by horizontal reflection and cropping~\cite{krizhevsky2012imagenet}, resulting in $0.8$M training images.
The \data{svhn} dataset is a large collection of digits ($10$ classes) cropped from real-world scenes, consisting of $73257$ training images, $26032$ testing images and $531131$ less difficult images for additional training.
In our experiments, we use all available training images, for a total of $604388$ training samples.
For the \data{mnist} variants~\cite{larochelle2007empirical}, each variation either reduces the training size ({\sc{mnist-07}}) or amends the original digits by rotation ({\sc{rot}}), background superimposition ({\sc{bg-rand}} and {\sc{bg-img}}), or a combination thereof ({\sc{bg-rot}}). 
We preprocess all datasets with whitening (except \data{cifar100} and \data{svhn} which were prohibitively large).

\paragraph{Baselines.}
We compare the proposed \abbrev{} with four baseline methods:
HashedNets~\cite{chen2015compressing}, low-rank decomposition (LRD)~\cite{denil2013predicting}, filter dropping (DropFilt) and frequency dropping (DropFreq).
HashedNets was first proposed to compress fully-connected layers in deep neural networks via the hashing trick.
In this baseline, we apply the hashing trick directly to the convolutional layer by hashing filter weights in the spatial domain.
This induces random weight sharing across all filters in a single convolutional layer.
Additionally, we compare against low-rank decomposition of the convolutional filters~\cite{denil2013predicting}.
Following the method in~\cite{denton2014exploiting}, we unfold the four-dimensional filter tensor to form a two dimensional matrix on which we apply the low-rank decomposition.
The parameters of the decomposition are fine-tuned via back-propagation.
DropFreq learns parameters in the DCT frequency domain but sets high frequency components to $0$ to meet the compression requirement.
DropFilt compresses simply by reducing the number of filters in each convolutional layer.

All methods were implemented using Torch7 \cite{collobert2011torch7} and run on NVIDIA GTX TITAN graphics cards with $2688$ cores and $6$GB of global memory.
Model parameters are stored and updated as $32$ bit floating-point values.%
\footnote{The compression rates of all methods could be further improved by learning and storing parameters in lower precision~\cite{courbariaux2015low,gupta2015deep}.}

%

\begin{table}[t]
\centering 
\resizebox{0.9\linewidth}{!}{
\begin{tabular}{cc|ccccc|c}
	\!\textbf{Layer}\!& \textbf{Operation} 	&\!\textbf{Input dim.}\!&\!\textbf{Inputs}\!&\!\textbf{Outputs}\!&\!\textbf{C size}\!&\!\textbf{MP size}\!&\!\textbf{Parameters}\!\\
\hline
$1$ & C,RL       & $32\!\times\! 32$	& $3$    & $32$     & $5\!\times\! 5$ &                     & $2K$      \\
$2$ &\!C,MP,DO,RL\!& $32\!\times\! 32$	& $32$   & $64$     & $5\!\times\! 5$ & $2\!\times\! 2 (2)$ & $51K$     \\
$3$ & C,RL       & $16\!\times\! 16$	& $64$   & $64$     & $5\!\times\! 5$ &                     & $102K$    \\
$4$ &\!C,MP,DO,RL\!& $16\!\times\! 16$	& $64$   & $128$    & $5\!\times\! 5$ & $2\!\times\! 2(2)$  & $205K$    \\
$5$ &\!C,MP,DO,RL\!& $8\!\times\! 8$	  & $128$  & $256$    & $5\!\times\! 5$ & $2\!\times\! 2(2)$  & $819K$    \\
$6$ & FC,Softmax & $-$                & $4096$ & $10/100$ &                 &                     & $40/400K$ \\
\end{tabular}
}
\vspace{0.25em}
\caption{Network architecture. C: Convolution. RL: ReLu. MP: Max-pooling. DO: Dropout. FC: Fully-connected. The number of parameters in the fully-connected layer is specific to $32\!\times\! 32$ input images and varies with the number of classes, either $10$ or $100$ depending on the dataset.}
\label{tab:architecture}
\end{table}


\begin{table}
\centering
\resizebox{\linewidth}{!}{
\begin{tabular}{r||c|ccccc||c|ccc}
	& \multicolumn{6}{c||}{\textbf{(a) Compression} $\mathbf{\!=\!1/16}$} & \multicolumn{4}{c}{\textbf{(b) Compression} $\mathbf{\!=\!1/64}$} \\
	&\!\textbf{CNN}\!&\!\!\textbf{DropFilt}\!\!&\!\!\textbf{DropFreq}\!\!&\!\textbf{LRD}\!&\!\!\textbf{HashedNets}\!\!&\!\!\textbf{\abbrevtable{}}\!\!&\!\textbf{CNN}\!&\!\textbf{LRD}\!&\!\!\textbf{HashedNets}\!\!&\!\!\textbf{\abbrevtable{}\!\!} \\
\hline
\!\data{cifar10}	\!&\! $14.91$ \!&\! $54.87$ \!&\! $30.45$ \!&\! $23.23$ \!&\! $24.70$ \!&\! \boldBlue{21.42} \!&\! $14.37$ \!&\! $34.35$ \!&\! $43.08$ \!&\! \boldBlue{30.79} \!\\
\!\data{cifar100}	\!&\! $33.66$ \!&\! $81.17$ \!&\! $55.93$ \!&\! $51.88$ \!&\! $48.64$ \!&\! \boldBlue{47.49} \!&\! $33.76$ \!&\! $66.44$ \!&\! $67.06$ \!&\! \boldBlue{62.33} \!\\
\!\data{svhn}			\!&\! $3.71$  \!&\! $30.93$ \!&\! $14.96$ \!&\! $10.67$ \!&\! $9.00$  \!&\! \boldBlue{8.01}  \!&\! $3.69$  \!&\! $22.32$ \!&\! $23.31$ \!&\! \boldBlue{18.37} \!\\
\!\data{mnist-07}	\!&\! $0.80$  \!&\! $4.90$  \!&\! $2.20$  \!&\! $1.18$  \!&\! $1.10$  \!&\! \boldBlue{0.94}  \!&\! $0.85$  \!&\! $1.95$  \!&\! $1.77$  \!&\! \boldBlue{1.24}  \!\\
\!\data{rot}			\!&\! $3.42$  \!&\! $29.74$ \!&\! $8.39$  \!&\! $4.79$  \!&\! $5.53$  \!&\! \boldBlue{3.87}  \!&\! $3.32$  \!&\! $9.90$  \!&\! $10.10$ \!&\! \boldBlue{6.60}  \!\\
\!\data{bg-rot}		\!&\! $11.42$ \!&\! $88.88$ \!&\! $56.63$ \!&\! $20.19$ \!&\! \boldBlue{16.15} \!&\! $18.43$ \!&\! $11.28$ \!&\! $35.64$ \!&\! $32.40$ \!&\! \boldBlue{27.91} \!\\
\!\data{bg-rand}	\!&\! $2.17$  \!&\! $90.10$ \!&\! $8.83$  \!&\! $2.94$  \!&\! $2.80$  \!&\! \boldBlue{2.63}  \!&\! $1.77$  \!&\! $4.57$  \!&\! $5.10$  \!&\! \boldBlue{3.62}  \!\\
\!\data{bg-img}		\!&\! $2.61$  \!&\! $89.41$ \!&\! $27.89$ \!&\! $4.35$  \!&\! \boldBlue{3.26}  \!&\! $3.97$  \!&\! $2.38$  \!&\! $7.23$  \!&\! \boldBlue{6.68}  \!&\! $8.04$  \!\\
\end{tabular}
}
\caption{Test error rates (in $\%$) with compression factors $1/16$ and $1/64$. Convolutional layers were compressed by the indicated methods (DropFilt, DropFreq, LRD, HashedNets, and \abbrevtable{}), with no \emph{convolutional layer} compression applied to CNN. The \emph{fully connected} layer is compressed by HashNets for \emph{all methods}, including CNN.}
\label{table:nlcf46}
\end{table}

\paragraph{Comprehensive evaluation.}
We adopt the network network architecture shown in Table~\ref{tab:architecture} for all datasets.
The architecture is a deep convolutional neural network consisting of five convolutional layers (with $5\times 5$ filters) and one fully-connected layer.
Before convolution, input feature maps are zero-padded such that output maps remain the same size as the (un-padded) input maps after convolution.
Max-pooling is performed after convolutions in layers $2$, $4$ and $5$ with filter size $2\times 2$ and stride $2$, reducing both input map dimensions by half.
Rectified linear units are adopted as the activation function throughout.
The output of the network is a softmax function over labels.

\begin{wrapfigure}{r}{0.63\textwidth}
\vspace{-1ex}
    \includegraphics[width=0.63\textwidth]{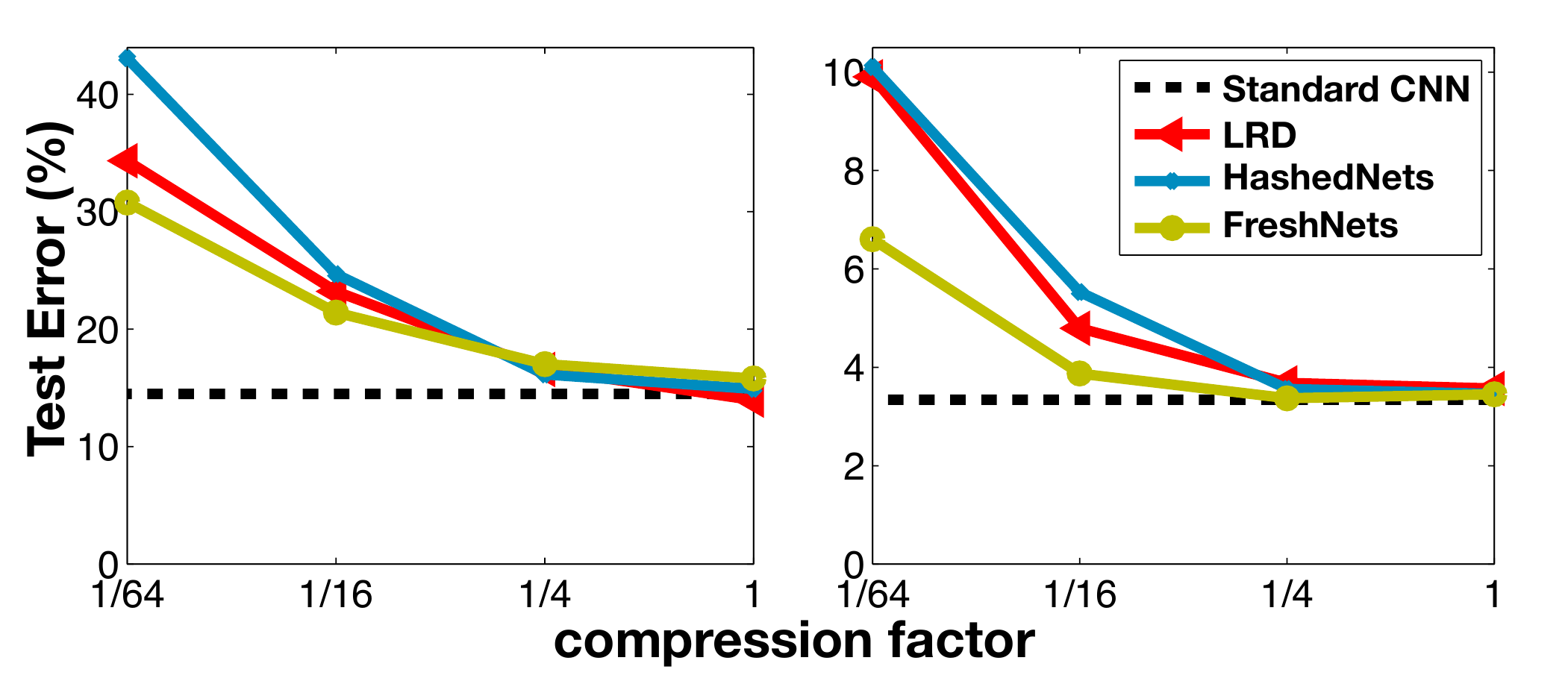}
    \vspace{-4ex}
	\caption{Test error rates at varying compression levels for datasets \data{cifar10} (left) and \data{rot} (right).}
    \label{fig:trace}
\end{wrapfigure}
In this architecture, the convolutional layers hold the majority of parameters ($1.2$ million in convolutional layer \emph{v.s.} $40$ thousand in the fully connected layer with $10$ output classes).
During training, we optimize parameters using mini-batch gradient descent with batch size $64$ and momentum $0.9$.
We use $20$ percent of the training set as a validation set for early stopping.
For \abbrev{}, we use a frequency-sensitive compression scheme which increases weight sharing among higher frequency components.\footnote{We evaluate several frequency-sensitive schemes later in this section, but for this comprehensive evaluation we set frequency compression rates by a rescaled beta distribution with $\alpha=0.25$ and $\beta=2.5$ for all layers.}
For all baselines, we apply HashedNets~\cite{chen2015compressing} to the fully connected layer at the corresponding level of compression.
All error results are reported on the test set.

Table~\ref{table:nlcf46}(a) and (b) show the comprehensive evaluation of all methods under compression ratios $1/16$ and $1/64$, respectively.
We exclude DropFilt and DropFreq in Table~\ref{table:nlcf46}(b) because neither supports $1/64$ compression in this architecture for all layers.
For all methods, the fully connected layer (top layer) is compressed by HashedNets~\cite{chen2015compressing} at the corresponding compression rate.
In this way, the final size of the entire network respects the specified compression ratio.
For reference, we also show the error rate of a standard convolutional neural network (CNN, columns 2 and 8) with the fully-connected layer compressed by HashedNets and \emph{no compression} in the convolutional layers.
Excluding this reference, we highlight the method with best test error on each dataset in \boldBlue{\textrm{\textbf{bold}}}.

\begin{wrapfigure}{r}{0.5\textwidth}
\vspace{-2ex}
    \includegraphics[width=0.5\textwidth]{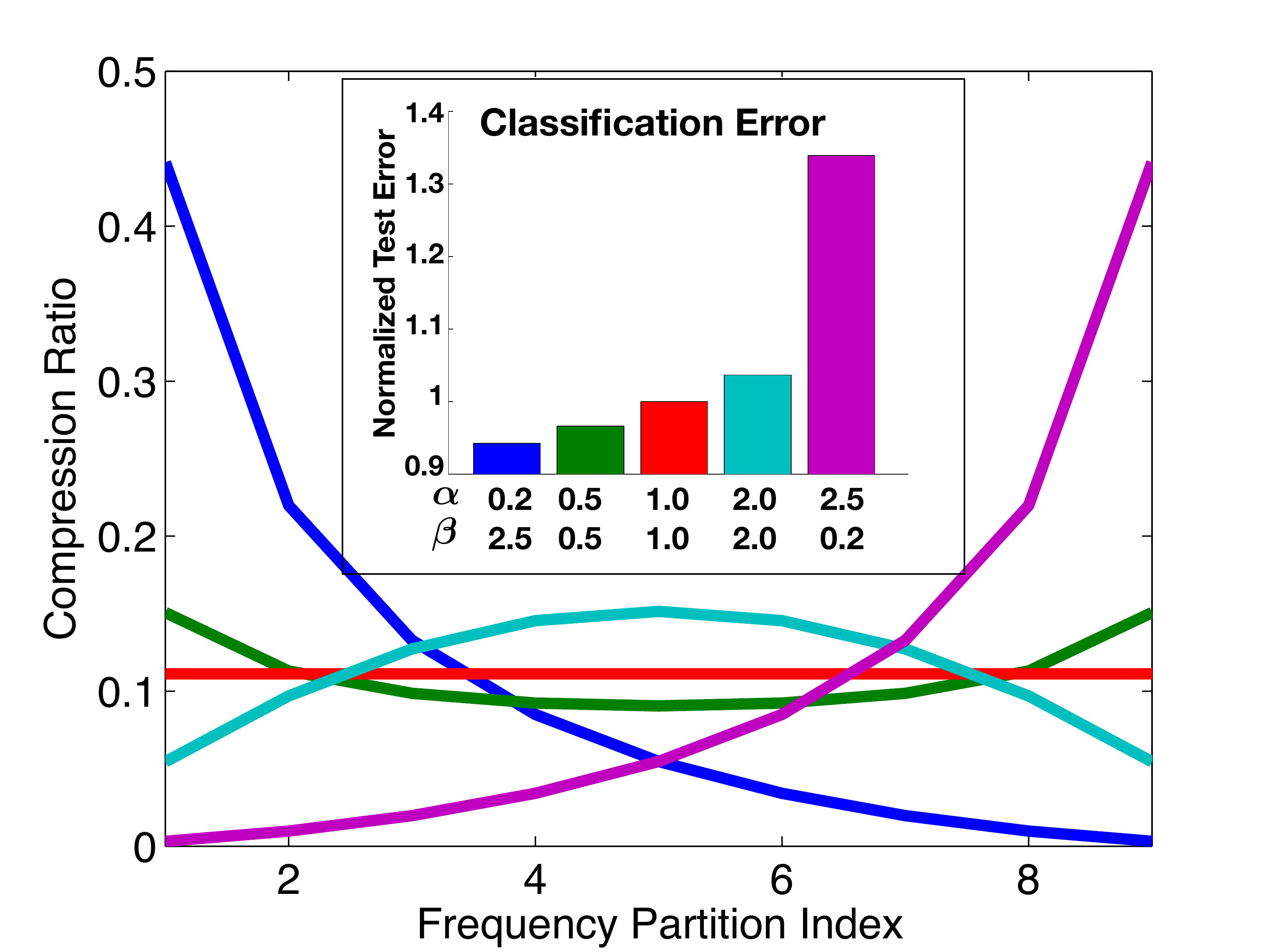}
    \vspace{-4ex}\caption{Results with different frequency sensitive compression schemes, each adopting a different beta distribution as the compression rate for each frequency. The inner figure shows normalized test error of each scheme on \data{cifar10} with the beta distribution hyper-parameters. The outer figure depicts the five beta distributions (with colors matching the inner figure).}
    \label{fig:beta_hashing}
\end{wrapfigure}
We discern several general trends.
In Table~\ref{table:nlcf46}(a), we observe the performance of the DropFilt and DropFreq at $1/16$ compression.
At this compression rate, DropFilt corresponds to a network $1/16$ filters at each layer: $2$, $4$, $4$, $8$, $16$ at layers $1\!-\!5$ respectively.
This architecture yields particularly poor test accuracy, including essentially random predictions on three datasets.
DropFreq, which at $1/16$ compression parameterizes each filter in the original network by only $1$ or $2$ low-frequency values in the DCT frequency space, performs with similarly poor accuracy.
Low rank decomposition (LRD) and HashedNets each yield similar performance at both $1/16$ and $1/64$ compression.
Neither explicitly considers the smoothness inherent in learned convolutional filters, instead compressing the filters in the spatial domain.
Our method, \abbrev{}, consistently outperforms all baselines, particularly at the higher compression rate as shown in Table~\ref{table:nlcf46}(b).
Using the same model in Table~\ref{tab:architecture}, Figure~\ref{fig:trace} shows more complete curves of test errors with multiple compression factors on the \data{cifar10} and \data{rot} datasets.

\paragraph{Varying compression by frequency.}
As mentioned in Section~\ref{subsec.hierarchical_hashing}, we allow a higher collision rate in the high frequency components than in the low frequency components for each filter.
To demonstrate the utility of this scheme, we evaluate several hash compression schemes.
Systematically, we set the compression rate of the $j^{th}$ frequency band $r_j$ with a parameterized function, \ie{} $r_j=f(j)$.
In this experiment, we use the beta distribution:
	$f(j;\alpha,\beta) = Z x^{\alpha-1} (1-x)^{\beta-1}$,
where $x\!=\!\frac{j+1}{2k-1}$ is a real number between 0 and 1, $k$ is the filter size, and $Z$ is a normalizing factor such that the resulting distribution of parameters meets the target parameter budget $K$, \ie{} $\sum_{j=0}^{2k-2} r_j N_j = K$.
We adjust $\alpha$ and $\beta$ to control the compression rate for each frequency region.
As shown in Figure~\ref{fig:beta_hashing}, we have multiple pairs of $\alpha$ and $\beta$, each of which results in a different compression scheme. For example, if $\alpha=0.25$ and $\beta=2.5$, the compression rate monotonically decreases as a function of component frequency, meaning more parameter sharing among high frequency components (blue curve in Figure~\ref{fig:beta_hashing}).

To quickly evaluate the performance of each scheme, we use a simple four-layer \abbrev{} where the first two layers are DCT-hashed convolutional layers (with $5\times 5$ filters) containing $32$ and $64$ feature maps respectively, and the last two layers are fully connected layers.
We test \abbrev{} on \data{cifar10} with each of the compression schemes shown in Figure~\ref{fig:beta_hashing}.
In each, weight sharing is limited to be within groups of similar frequencies, as described in Section~\ref{subsec.hierarchical_hashing}, however number of unique weights shared within each group is varied.
We denote the compression scheme with $\alpha,\beta=1$ (red curve) as a \emph{frequency-oblivious scheme} since it produces a uniform compression independent of frequency.
In the inset bar plot in Figure~\ref{fig:beta_hashing}, we report test error normalized by the test error of the frequency-oblivious scheme and averaged over compression rates $1$, $1/2$, $1/4$, $1/16$, $1/64$, and $1/256$.
We can see that the proposed scheme with fewer shared weights allocated to high frequency components (represented by the blue curve) outperforms all other compression schemes.
An inverse scheme where the high frequency regions have the lowest collision rate (purple curve) performs the worst.
These empirical results fit our assumption that the low frequency components of a filter are more important than the high frequency components.

\begin{figure}
	\center
    \includegraphics[width=1\textwidth]{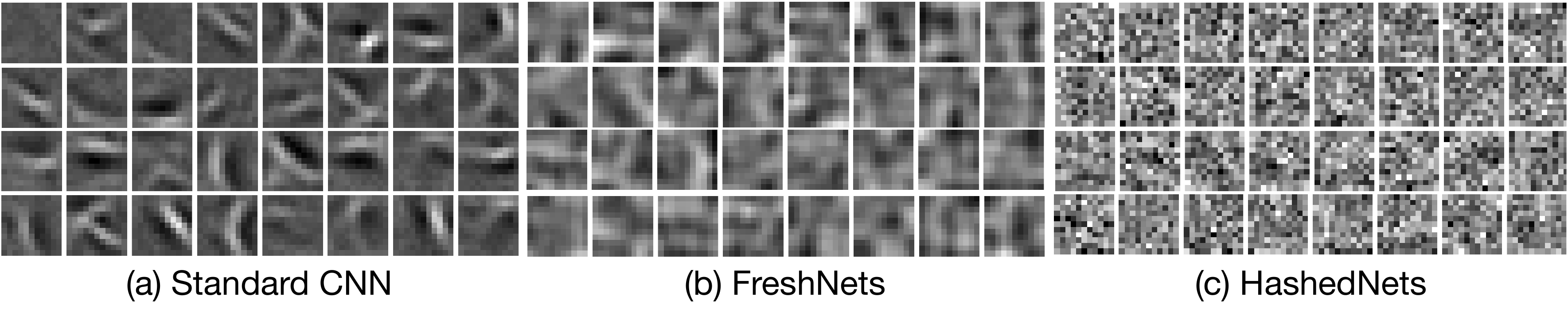} 
    \caption{Visualization of filters learning on \data{mnist} in (a) an uncompressed CNN, (b) a CNN compressed with \abbrev{}, and (c) a CNN compressed with HashedNets (compression rate $1/16$ in both (b) and (c)). \abbrev{} preserves the smoothness of the filters, whereas  HashedNets does not.}
    \label{fig:visual}
\end{figure}

\paragraph{Filter visualization.}
We investigate the smoothness of the learned convolutional filters in Figure~\ref{fig:visual} by visualizing the filter weights (first layer) of (a) a standard, uncompressed CNN, (b) \abbrev{}, and (c) HashedNets (with weight sharing in the spatial domain).
For this experiment, we again apply a four layer network with two convolutional layers but adopt larger filters ($11\times 11$) for better visualization.
All three networks are trained on \data{mnist}, and both \abbrev{} and HashedNets have $1/16$ compression on the first convolutional layer.
When plotting, we scale the values in each filter matrix to the range $[0,255]$.
Hence, white and black pixels stand for large positive and negative weights, respectively.
We observe that, although more blurry due to the compression, the filter weights of \abbrev{} are still smooth while weights in HashedNets appear more chaotic.

\section{Conclusion}
In this paper we present \abbrev{}, a method for learning convolutional neural networks with dramatically compressed model storage.
Harnessing the hashing trick for parameter-free random weight sharing and leveraging the smoothness inherent in convolutional filters, \abbrev{} compresses parameters in a frequency-sensitive fashion such that significant model parameters (\emph{e.g.} low-frequency components) are better preserved.
As such, \abbrev{} preserves prediction accuracy significantly better than competing baselines at high compression rates.


{
\small
\bibliographystyle{ieee}
\bibliography{hashnn,kilian,wenlin}
}

\end{document}